*Technical Note*

# Image Reconstruction of Multi Branch Feature Multiplexing Fusion Network with Mixed Multi-layer Attention

Yuxi Cai [1], Guxue Gao [2] Zhenhong Jia [2] and Huicheng Lai*

College of Information Science and Engineering, Xinjiang University, Urumqi 830000, China;
cai@stu.xju.edu.cn (Y.C.); gaoyangshuang123@stu.xju.edu.cn (G.G.); jzhh@xju.edu.cn (Z.J.)
*Correspondence: lai@xju.edu.cn

**Abstract:** Image super-resolution reconstruction achieves better results than traditional methods with the help of the powerful nonlinear representation ability of convolution neural network. However, some existing algorithms also have some problems, such as insufficient utilization of phased features, ignoring the importance of early phased feature fusion to improve network performance, and network can't pay more attention to high-frequency information in the reconstruction process. To solve these problems, we propose a multi branch feature multiplexing fusion network with mixed multi-layer attention (MBMFN), which realizes the multiple utilization of features and the multi-stage fusion of different levels of features. To further improve the network's performance, we propose a lightweight enhanced residual channel attention (LERCA), which can not only effectively avoid the loss of channel information, but also make network pay more attention to the key channel information and benefit from it. Finally, the attention mechanism is introduced into the reconstruction process to strengthen the restoration of edge texture and other details. A large number of experiments on several benchmark sets show that compared with other advanced reconstruction algorithms, our algorithm produces highly competitive objective indicators and restores more image detail texture information.

## 1. Introduction

For a long time, single image super-resolution reconstruction (SISR) is a classic research problem in the field of computer bottom vision. It aims to use certain technical means to restore the corresponding high-resolution image containing rich texture information from the degraded low-resolution image. SISR has been widely used in remote sensing, public security, medicine and other fields. There are many different high-resolution images in the real scene, after different degradation, the same low-resolution image is obtained. Therefore, SISR is a typical ill posed problem. Benefiting from the vigorous development of machine learning, researchers have proposed many reconstruction algorithms based on deep learning, DRN [1], HAN [2], and LatticeNet [3], and achieved better results than traditional algorithms. DRN constructs a double regression network, which provides degradation information constraints for the network by modeling the real image degradation process. HAN designs layer attention module and channel space attention module. The network models the feature information between layers, channels and locations through the two attention modules, which makes the network achieve good results. LatticeNet designs a lattice filter based on butterfly structure and integrates context information through reverse feature fusion strategy, at the expense of computational complexity and memory storage, the network achieves good results.

Dong et al. [4] constructed a shallow end-to-end nonlinear mapping neural network for the first time, and achieved better results than traditional algorithms such as interpolation. Subsequently, researchers have built a series of very deep neural networks, but with the deepening of the number of network layers and the further improvement of the reconstruction effect, they are also faced with the problem that the network is difficult to train. Kim et al. [5] constructed VDSR by introducing residual learning. With the help of jump connection, the network was further deepened and the performance of model was improved again. Inspired by this, Lim et al. [6] removed the commonly used batch normalization layer in the network and constructed EDSR using small-scale residual blocks, which not only saved the storage cost, but also improved the reconstruction effect. Liu et al. [7] believe that the features after residual learning will form more complex fusion features, which makes the network ignore the cleaner residual features generated in the middle. Therefore, Liu transmits the residual features generated in the middle to the end of the basic block for local fusion through jump connection, which greatly improves the reconstruction effect.

Most CNN based algorithms have achieved excellent results, but they still face some problems. Usually, most of the extracted phased intermediate features will be processed by a series of stacked convolution layers, and the generated intermediate features are rarely used, or only used by the network once, they can't be further processed by the network, resulting in a certain degree of feature waste. Then, the features transferred to the next layer are processed by convolution layer to form new features more complex than the original features. Liu et al. [8] have shown that the same feature will show different information in different feature extraction stages of network. At the same time, due to different receptive fields, the same feature will also be extracted with different information, which contributes to the reconstruction results to varying degrees. Most algorithms focus on the final generated features and ignore the reuse of intermediate features, which leads to the decline of network performance to a certain extent. In addition, another little attention is paid to the stage feature fusion early. Most reconstruction algorithms only fuse the features in different stages at the end of the basic block to form complex features with relatively rich high-frequency information. However, the feature fusion in different stages in the early stage can not only aggregate the hierarchical information under different sensory fields, but also further deepen the fusion features, it indirectly realizes the extraction and fusion of multi-scale features, and further expands the receptive field of the network. Finally, due to the use of deconvolution to realize up sampling, it will bring different degrees of artifacts to the reconstructed image and affect the reconstruction effect. Most reconstruction algorithms use sub-pixel convolution [9] to realize up sampling of feature size. In the reconstruction process, these algorithms can't make the network effectively focus on high-frequency information such as local edge texture, resulting in poor recovery effect.

To solve the above problems, we design MBMFN, which realizes the multiple utilization of features and multi-stage local feature fusion, so that the network can learn a more discriminative feature representation. Meanwhile, with the help of the attention mechanism in the reconstruction process, we can better recover the edge texture and other details of the image. Our contribution mainly includes the following points:

1 A multi branch feature reuse fusion attention block is proposed, which realizes the reuse of features and the fusion of multi-stage local features through the interaction of information between multiple branches, and enriches the hierarchical types of phased features.

2 A lightweight enhanced residual channel attention (LERCA) is proposed, which can pay more attention to the high-frequency information in low resolution space. We use 1×1 convolution to establish the interdependence between channels, which avoids the loss of some channel information caused by channel compression, and the module is more lightweight.

3 In the reconstruction process, the attention mechanism is introduced, and the U-LERCA is constructed combined with LERCA, which enhances the sensitivity of the network to key information. In particular, few people have studied the attention strategies in the reconstruction process.

4 We construct a multi-branch feature multiplexing fusion network with mixed multi-layer attention, which achieves a good recovery effect. At the same time, the network achieves a good balance between parameters and performance.

**2. Related Work**

2.1 Phased Feature Fusion

With the vigorous development of deep learning, many lightweight reconstruction algorithms have been proposed, and achieved good results in objective indicators and subjective vision. In order to strengthen the fusion of different levels of features, Hui et al. introduced the idea of distillation in IDN [10], and divided the features processed by convolution layer into two parts through channel splitting operation. Part of the features continue to be further processed by convolution layer, and the remaining features are spliced with the original input features and transmitted to the end of the enhancement module by jumping for cross fusion of different local features, in order to strengthen network learning of LR contour region. On this basis, Hui continues to deeply study the application of distillation idea and local feature fusion, and put forward IMDN [11]. The network uses channel splitting strategy to distill fine features layer by layer, aggregates different levels of features through splicing and 1×1 convolution layer, but uses channel splitting operation for feature distillation, which brings a certain degree of inflexibility to the network. To this end, Liu et al. proposed RFDN [12], which uses 1×1 convolution layer instead of channel splitting operation to carry out compressed feature distillation, while using convolution layer with residual to replace the original convolution layer, the network becomes lighter and the performance is further improved, but the network only extracts feature at a fixed scale and can't effectively aggregate feature information of different scales. Wang et al. proposed MSFIN [13], which restores the input image to the target size by interpolation algorithm, uses 3×3 convolution layer to down-sampling different image sizes, and then uses three branches to process the features of different image sizes in parallel. In order to make up for the lack of information exchange between branches, deconvolution is used to restore the feature size, and the up-sampled features are fused with the features of the corresponding stage of the previous branch. Through this design, the network effectively integrates features of different sizes and benefits from it, but the network runs in high-resolution space, resulting in a lot of memory and computational overhead.

2.2 Attention Mechanism

Attention mechanism can effectively guide neural network to focus on the most important information in the input features and strengthen the network's learning and expression of these information. It has been widely used in various computer vision tasks, including image segmentation, target tracking, image restoration and so on, and has shown great advantages in improving the performance of network. Hu et al. introduced the attention mechanism to the image classification task and proposed SENet [14] network which explicitly models the interdependence between channels, which can adaptively correct channel features and retain valuable features, so the network performance is further improved. Roy et al. [15] use 1×1 convolution layer instead of full connection layer in SENet to model the dependence between channels. Wang et al. [16] generate channel weights through lighter and faster one-dimensional convolution. Recently, Hui et al. [11] proposed Contrast-aware channel attention, and the network not only achieves a good objective index, but also restores more detailed information such as edge texture. Niu et al. [17] constructed a hybrid attention mechanism to adaptively capture key information. Liu et al. [18] further enhance the receptive field of spatial attention with the help of convolution layer and pooling layer, so that the network can learn more context information.

## 3 Paper Method

In this part, we first introduce the multi branch feature multiplexing fusion network proposed in this paper, and then introduce each component of the network in detail.

3.1 Overall Framework

As shown in figure 1, the multi-branch feature multiplexing fusion network with mixed multi-layer attention (MBMFN) is composed of shallow feature extraction block, multi-branch feature multiplexing fusion attention block (MBMFB) and reconstruction block. In this paper, $I_{LR}$ $I_{SR}$ are represented as the input and output images of the network respectively. According to the research of [19] and [20], in this paper, only a 3×3 convolution layer is used for shallow feature extraction of $I_{LR}$:

$$F_0 = Conv_{3\times3}(I_{LR}) \qquad (1)$$

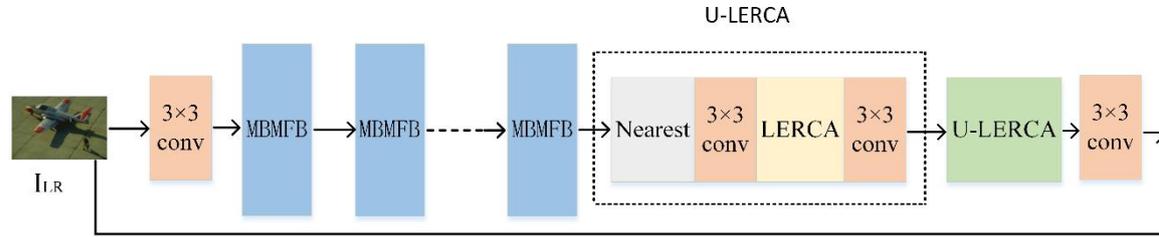

**Fig. 1** overall block diagram of multi-branch feature multiplexing fusion network with mixed multi-layer attention

Where $Conv_{3\times3}(\cdot)$ represents convolution operation with convolution kernel of 3×3, and $F_0$ represents the shallow features extracted by the convolution layer. Then, the $F_0$ is used as the input of the MBMFB to deepen the features in order to learn a more discriminative feature representation.

Assuming that there are d MBMFBs, the output features of the d-th MBMFB are expressed as:

$$F_d = H_{MBMFB,d}(F_{d-1}) = H_{MBMFB,d}(H_{MBMFB,d-1}(\cdots(H_{MBMFB,1}(F_0))\cdots)) \qquad (2)$$

Where $H_{MBMFB,d}(\cdot)$ represents the d-th MBMFB with composite function. $F_d$ represents the local fusion feature extracted after the d-th MBMFB processing. More details about MBMFB will be described in detail in 3.2.

After multiple MBMFBs processing, we send the learned discriminative features into the reconstruction module with attention, so as to restore to the corresponding target size. The operation process is expressed as follows:

$$F_n = H_{U-LERCA}^n(H_{U-LERCA}^{n-1}(\cdots H_{U-LERCA}^0(F_d)\cdots)) \qquad (3)$$

Where $H_{U-LERCA}^n$ represents the n-th U-LERCA block, $F_n$ represents the output features of the n-th U-LERCA, and more information about U-LERCA will be introduced in 3.4.

In order to make up for the problem of losing part of the underlying information in the continuous deepening processing of features, we use the traditional interpolation algorithm to sample the $I_{LR}$ to the corresponding size, and supplement the information by jumping connection, so as to generate the final $I_{SR}$:

$$I_{SR} = \text{Conv}_{3\times3}(F_n) + H_{up}(I_{LR}) \tag{4}$$

Where, $H_{up}(\cdot)$ represents the up-sampling operation of bilinear interpolation.

According to the previous research work, we use $L_1$ loss function to optimize the network parameters. Given a training set $\{I_{LR}^j, I_{HR}^j\}_{j=1}^{j=N}$ that contains many image pairs, where N represent the number of training image pairs, the $L_1$ loss function with parameters used in this paper is expressed as follows:

$$L(\theta) = \frac{1}{N}\sum_{i}^{N}\left\|H_{\text{MBMFN}}(I_{LR}^j) - I_{HR}^j\right\|_1 \tag{5}$$

Here θ represents network parameters that need to be optimized, and $H_{\text{MBMFN}}(\cdot)$ represents the multi-branch feature multiplexing fusion network with mixed multi-layer attention.

3.2 Multi-branch Feature Multiplexing Fusion Attention Block

In order to realize the reuse of local features and promote the fusion of multi-level features, a semi-parallel multi-branch feature reuse fusion attention block is designed in this paper, which not only expands the receiving domain of network, but also avoids the deepening of network, the internal structure of the block is shown in figure 2.

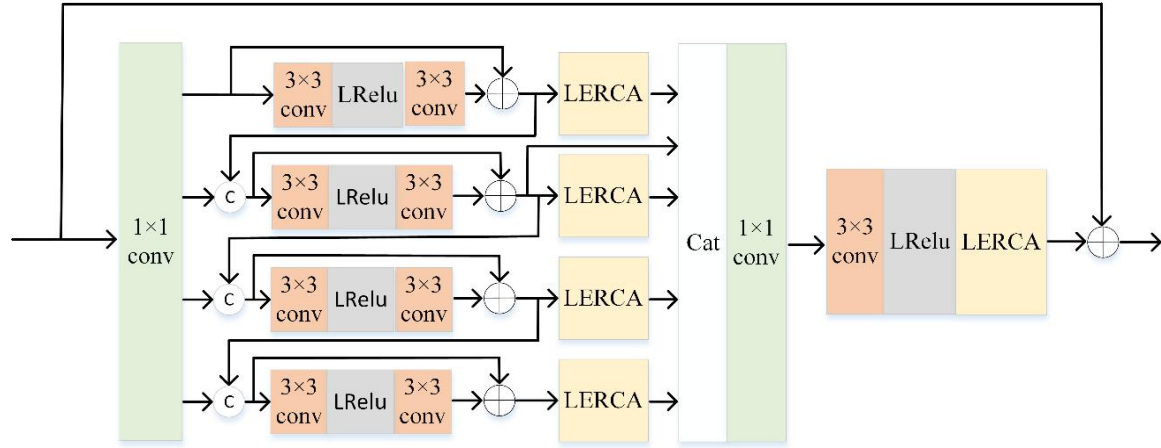

**Fig. 2** Internal structure diagram of multi-branch feature multiplexing fusion attention block

First of all, we use 1×1 convolution layer for feature distillation to reduce the amount of computation and redundant feature information of network, and then the refined features are sent to the four branches of parallel processing for feature extraction and fusion.

Each branch contains a residual block and LERCA. With the help of the residual block composed of two 3×3 convolution layers, the features after distillation are further deepened. In order to enhance the feature extraction ability of network and focus the network on the key features, we design LERCA in order to enhance the network's learning of key features, more details on the LERCA will be introduced in 3.3. We add LERCA to each branch to strengthen the network to extract the key information of hierarchical features in different channels, so that the network can learn more discriminative feature representation.

The output features after the residual block in the first branch and the previous distillation features are spliced together as the input of the second branch. With the help of the convolution layer of the second branch, the network integrates the hierarchical features of different stages, realizes the reuse of features and expands the acceptance domain of the network. Therefore, the 3×3 convolution layer shoulders the important task of feature fusion and extraction. Because the input of the second branch is spliced by the features of two stages, with the help of jump connection, the network adds the features of different stages to the same stage features, and further forms rich hierarchical features. By analogy, through feature reuse and feature processing of three or four branches, the network effectively integrates the features of multiple stages and forms more abundant hierarchical features.

Although channel attention can enhance the network's attention to information-rich channels, there is some key information in those suppressed channels. In order to make up for the loss of this information, we take the output feature after the residual block in the second branch as the basic bottom feature (as shown by the blue line in figure 2), and splice together with the output features of the other four branches in a local form. Then, 1×1 convolution layer is used for feature aggregation. Finally, the attention mechanism is used to strengthen the learning of important features in order to improve the representation ability of the network. In addition, at the end of MBMFB, we use residual connections to jump forward the original information, so as to benefit from residual learning and speed up the back propagation of gradient in network optimization.

3.3 Lightweight Enhanced Residual Channel Attention

As shown in Figure 3, although RCAB [21] proposed by Zhang et al. can adaptively adjust channel features response according to the interdependence between channels, and achieve quite good performance, RCAB compresses the channel, resulting in varying degrees of channel information loss. Secondly, RCAB uses 3×3 convolution layer with a large number of parameters for many times, which brings a lot of memory overhead to the network and is not suitable for lightweight network reconstruction. Therefore, LERCA is proposed in this paper.

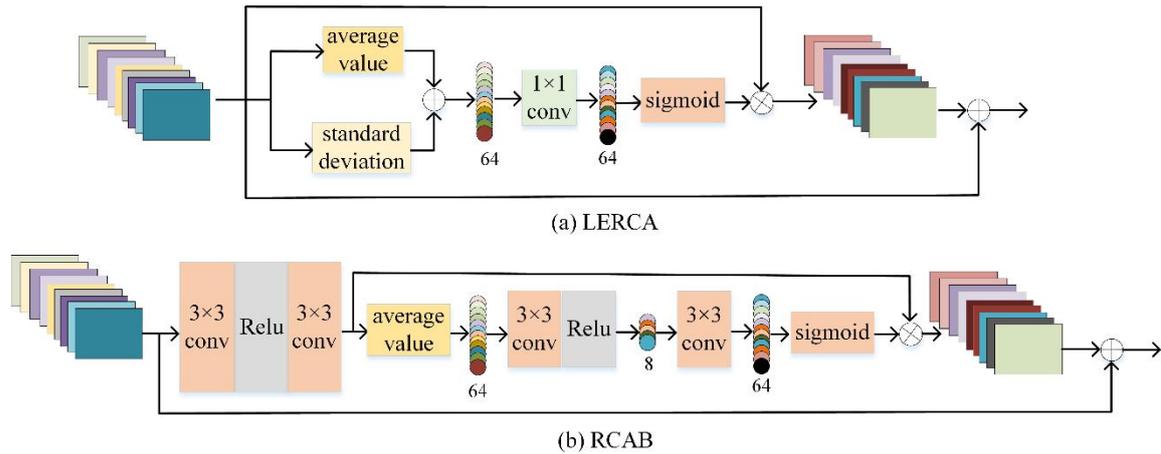

**Figure 3** (a) lightweight enhanced residual channel attention (b) residual channel attention block

The LERCA proposed in this paper differs from RCAB in the following three ways. First, we removed the two convolution layers in front of RCAB. Behjati et al. [22] think that channel attention may discard some relevant detailed features, which will be difficult to regain at a deeper network level. At the same time, the 3×3 convolution layer will bring many parameters to the network. In order to supplement the relevant detailed features and meet the needs of the lightweight network, we remove the previous two convolution layers. Due to the existence of convolution layer, it is difficult for RCAB to compensate for the missing details information through jump connection, but our LERCA can easily compensate the lost feature information. Second, we use the sum of global average and standard deviation instead of global average pooling. Compared with the global average pooling, the sum of the global average and standard deviation can more fully characterize the characteristics of the channel. Finally, 1×1 convolution layer is used to replace the two 3×3 convolution layers in RCAB. RCAB uses two 3×3 convolution layers for channel compression and recovery respectively, which destroys the channel characteristics and causes the loss of channel information to a certain extent. We directly use 1×1 convolution layer to model the interdependence between channels, meanwhile, reduces the number of parameters and achieves the goal of lightening the module.

In short, we use the sum of the global average and standard deviation of features to characterize the channel characteristics, directly model the relationship between channels with the help of 1 × 1 convolution layer, and then generate an attention mask through the Sigmoid function. The module can not only reduce the loss of channel information, but also help the network to recover more details such as image edge texture. With the help of jump connection, the network effectively makes up for the relevant information lost due to channel attention, and this information can no longer be regained in the deeper layers of the network.

3.4 Reconstruction block(U-LERCA)

In some previous reconstruction algorithms, the reconstruction block of network is usually composed of convolution layer and sub-pixel convolution layer, or the traditional interpolation algorithm is used to realize up-sampling operation, and the attention mechanism is rarely introduced into the reconstruction process. As a result, the key features are difficult to play their due role. On the other hand, for large-scale reconstruction tasks, such as ×4, if there is a lack of enough high-frequency information, it is difficult for network to achieve satisfactory reconstruction results. In view of the above two points, we combine LERCA to build U-LERCA, whose internal structure is shown in figure 1, which is composed of the nearest neighbor interpolation algorithm, LERCA and two convolution layers.

In U-LERCA, we abandon the sub-pixel convolution and use nearest neighbor interpolation algorithm to achieve up-sampling operation, mainly because the sub-pixel convolution layer brings many parameters to network, but also can't achieve the corresponding recovery effect. We use convolution layer to further establish the correlation between channels after interpolation. In order to distinguish the importance between channels and play the role of some key features, we use LERCA to enhance the network's attention to key channels to achieve the purpose of improving network performance.

In order to cope with the fact that under the large-scale reconstruction task, the network still has enough high-frequency information available, while ensuring that each step of the up-sampling is optimal, we adopt a distributed strategy to carry out step-by-step up-sampling until the target size is reached. For example, the ×4 reconstruction task is divided into two cascaded ×2 reconstruction tasks in this paper. At the same time, thanks to the addition of the attention mechanism in the reconstruction process, the network achieves a good reconstruction performance.

**4. Experiment and Analysis**

In this paper, the DIV2K [23] data set is used for network training. In the testing phase, four standard benchmark data sets are adopted: Set5, Set14, B100, Urban100. In this paper, the image is converted from RGB color space to YCbCr color space, and only Y channel is trained and tested. The Peak Signal-to-Noise Ratio (PSNR) and Structure Similarity Index Measure (SSIM) are used to quantitatively analyze the reconstruction results.

4.1 Experimental Environment and Parameter Setting

In the reconstruction tasks of ×2, ×3 and ×4, we randomly extract 24 image patches of 192×192 as the input of the network. One epoch is formed by 1000 back propagation iterations, and the initial learning rate is 0.0002. After each 200 epochs, learning rate decays to half of the original. 6 MBMFBs are used in the network, and the activation function Leaky-Relu is set to 0.05. the Adam algorithm is used to optimize the network gradient. Under the framework of Pytorch deep learning, we construct the MBMFN algorithm, the experimental hardware platform is NVIDIA Tesla V100-PCIE-16GB, and the software environment is Windows10 operating system.

4.2 Ablation Experiment

In order to verify effectiveness of the network design, we carried out ablation experiments on some of the important blocks. All ablation experiments were based on the training of 400 epochs under the ×4 reconstruction task.

Verify the impact of the location of information input between branches in MBMFB on network performance: as shown in Table 1, where Basic Branch refers to the branch in the second branch of MBMFB that skips LERCA, as shown in the blue line in figure 2, this branch is mainly used to supplement underlying information. Before Residual Way (BRW) refers to the input of features before residual to the next branch. After Residual Way (ARW) refers to the input of features after residual to the next branch. After Attention Way (AAW) refers to the input of features processed by LERCA to the next branch.

**Table 1** influence of information input location between branches in basic blocks on network performance

| Branch Input Type | BRW | ARW | AAW | PSNR(Set5) | SSIM(Set5) |
|---|---|---|---|---|---|
| Basic Branch | | | | | |
| | √ | × | × | 32.209 | 0.8940 |
| No | × | √ | × | 32.231 | 0.8945 |
| | × | × | √ | 32.216 | 0.8939 |
| | √ | × | × | 32.223 | 0.8943 |
| Yes | × | √ | × | 32.247 | 0.8945 |
| | × | × | √ | 32.243 | 0.8942 |

seen from Table 1, with or without Basic Branch, ARW has achieved better results than other input patterns, which may benefit from residual learning. In the case of no Basic Branch, the effect of AAW is significantly lower than that of ARW, which may be due to the fact that some of the key information in the suppressed channel features can't be effectively expressed after LERCA processing, resulting in performance degradation. Compared with the situation with Basic Branch, ARW has achieved a good improvement effect, which is due to the fact that Basic Branch supplement the neglected feature information and make up for the loss of underlying information caused by the attention mechanism.

Verify the validity of LERCA: in order to verify the effectiveness of LERCA, we replace LERCA in MBMFB with Squeeze-and-Excitation (SE), Channel attention (CA), Residual channel attention (RCA) and Contrast-aware channel attention (CCA) respectively, in which RCA only adds residual connections on the basis of CA. The experimental results are shown in Table 2.

**Table 2** performance comparison of various attention mechanisms in basic blocks

| Attention Type | MBMFB-No | MBMFB-SE | MBMFB-CA | MBMFB-RCA | MBMFB-CCA | MBMFB-LERCA |
| --- | --- | --- | --- | --- | --- | --- |
| PSNR(Set5) | 32.164 | 32.214 | 32.233 | 32.206 | 32.208 | 32.247 |
| SSIM(Set5) | 0.8937 | 0.8943 | 0.8942 | 0.8944 | 0.8941 | 0.8954 |

It can be clearly seen from table 2 that various attention mechanisms have significantly improved the performance of the network. Compared with MBMFB-CA, PSNR and SSIM of MBMFB-LERCA on set5 test set are increased by 0.014db and 0.0012 respectively, which is due to LERCA's uncompressed processing of the channel, reducing the loss of channel information and protecting channel characteristics. Compared with MBMFB-RCA, PSNR and SSIM of MBMFB-LERCA increased by 0.041db and 0.001 respectively on Set5, which may be related to the fact that LERCA uses the sum of channel global average and standard deviation to characterize channel characteristics. Compared with MBMFB-CCA, PSNR and SSIM of MBMFB-LERCA increased by 0.039db and 0.0013 respectively, indicating that LERCA is more effective than CCA.

Verify the effectiveness of adding the attention mechanism in the reconstruction phase: in order to prove that the introduction of attention mechanism in the reconstruction process has a good improvement effect on the network, we have carried out experimental verification. The experimental results are shown in Table 3, in which U-Nearest-LERCA-×4 means that in the reconstruction process, the nearest neighbor interpolation is used to realize the up-sampling operation and the LERCA is added, and the up sampling is directly ×4 without step-by-step processing. U-Nearest-×2-×2$^{\text{Weight sharing}}$ means that the nearest neighbor interpolation is used to realize the up-sampling operation in the reconstruction process, LERCA is not added, and the ×4 reconstruction task is divided into two cascaded ×2 reconstruction task, the two reconstruction blocks use the weight sharing strategy. U-subpixel refers to the up-sampling operation using subpixel convolution.

**Table 3** effectiveness of attention mechanism and distributed processing in the reconstruction phase

| Up-sampling pattern | U-Nearest-×4 | U-Nearest-LERCA-×4 | U-Nearest-×2-×2$^{\text{Weight sharing}}$ | U-Nearest-LERCA-×2-×2$^{\text{Weight sharing}}$ | U-Nearest-LERCA-×2-×2$^{\text{No Weight sharing}}$ | U-Subpixel |
| --- | --- | --- | --- | --- | --- | --- |
| PSNR(Set5) | 32.207 | 32.221 | 32.219 | 32.247 | 32.246 | 32.216 |
| SSIM(Set5) | 0.8941 | 0.8941 | 0.8942 | 0.8945 | 0.8945 | 0.8943 |
| Parameter | 1220K | 1224K | 1220K | 1224K | 1291K | 1250K |

red with the reconstruction process without attention mechanism, U-Nearest-LERCA-×4 has higher 0.014dB than U-Nearest-×4 on PSNR. Similarly, U-Nearest-LERCA-×2-×2$^{\text{Weight sharing}}$ has higher 0.028dB than U-Nearest-×2-×2$^{\text{Weight sharing}}$, and it also has higher 0.031dB than U-Subpixel. On SSIM, U-Nearest-LERCA-×2-×2$^{\text{Weight sharing}}$ is 0.0003 higher than U-Nearest-×2-×2$^{\text{Weight sharing}}$. This directly and effectively proves that the introduction of

LERCA attention mechanism in the reconstruction process can improve the reconstruction performance of the network.

Effectiveness of step-by-step processing in reconstruction phase: this shows the effectiveness of the distributed processing strategy in the reconstruction process, and the experimental results are shown in Table 3. Compared with the single-step processing, U-Nearest-×2-×2$^{\text{Weight sharing}}$ has higher 0.012dB than U-Nearest-×4 on PSNR, similarly, U-Nearest-LERCA-×2-×2$^{\text{Weight sharing}}$ has higher 0.026dB than U-Nearest-LERCA-×4. On SSIM, U-Nearest-LERCA-×2-×2$^{\text{Weight sharing}}$ is 0.0004 higher than U-Nearest-LERCA-×4. Obviously, the use of distributed processing strategy can ensure that each reconstruction task has enough high-frequency information available, thus improving the performance of the network.

In order to further reduce the network parameters, we use the weight sharing strategy for the reconstruction blocks in the reconstruction process. As can be seen from Table 3, on PSNR, U-Nearest-LERCA-×2-×2$^{\text{Weight sharing}}$ has higher 0.001dB than U-Nearest-LERCA-×2-×2$^{\text{No Weight sharing}}$, and the number of parameters is 67K less.

4.3 Comparison with Other Advanced Algorithms

In order to prove effectiveness of this network, we compare MBMFN with other advanced lightweight super-resolution reconstruction algorithms with up-sampling factors of ×2, ×3 and ×4. Include SRCNN [4], VDSR [5], DRCN [24], MemNet [25], CARN [26], IMDN [11], DNCL [27], FilterNeL [28], RFDN [12], CFSRCNN [29], SeaNet-baseline [30], SMSR [31], MSFIN [13], Cross-SRN [32], MRFN [33], MADNet-L$_F$ [34]. The experimental results are shown in Table 4. Except for the algorithm in this paper, the results of other algorithms come from published papers.

**Table 4** average PSNR/SSIM of BI degradation models × 2, × 3 and × 4, and the optimal results are shown in bold.

| Scale | Method | Year | Set5 | | Set14 | | B100 | | Urban100 | |
|---|---|---|---|---|---|---|---|---|---|---|
| | | | PSNR | SSIM | PSNR | SSIM | PSNR | SSIM | PSNR | SSIM |
| ×2 | SRCNN | 2016 | 36.66 | 0.9542 | 32.42 | 0.9063 | 31.36 | 0.8879 | 29.50 | 0.8946 |
| | VDSR | 2016 | 37.53 | 0.9587 | 33.03 | 0.9124 | 31.90 | 0.8960 | 30.76 | 0.9140 |
| | DRCN | 2016 | 37.63 | 0.9588 | 33.04 | 0.9118 | 31.85 | 0.8942 | 30.75 | 0.9133 |
| | MemNet | 2017 | 37.78 | 0.9597 | 33.28 | 0.9142 | 32.08 | 0.8978 | 31.31 | 0.9195 |
| | CARN | 2018 | 37.76 | 0.9590 | 33.52 | 0.9166 | 32.09 | 0.8978 | 31.92 | 0.9256 |
| | IMDN | 2019 | 38.00 | 0.9605 | 33.63 | 0.9177 | 32.19 | 0.8996 | 32.17 | 0.9283 |
| | DNCL | 2019 | 37.65 | 0.9599 | 33.18 | 0.9141 | 31.97 | 0.8971 | 30.89 | 0.9158 |
| | MRFN | 2019 | 37.98 | **0.9611** | 33.41 | 0.9159 | 32.14 | 0.8997 | 31.45 | 0.9221 |
| | FilterNeL | 2020 | 37.86 | 0.9610 | 33.34 | 0.9150 | 32.09 | 0.8990 | 31.24 | 0.9200 |
| | MADNet-L$_F$ | 2020 | 37.85 | 0.9600 | 33.39 | 0.9161 | 32.05 | 0.8981 | 31.59 | 0.9234 |
| | RFDN | 2020 | 38.05 | 0.9606 | 33.68 | 0.9184 | 32.16 | 0.8994 | 32.12 | 0.9278 |
| | CFSRCNN | 2020 | 37.79 | 0.9591 | 33.51 | 0.9165 | 32.11 | 0.8988 | 32.07 | 0.9273 |
| | SeaNet-baseline | 2020 | 37.99 | 0.9607 | 33.60 | 0.9174 | 32.18 | 0.8995 | 32.08 | 0.9276 |
| | SMSR | 2021 | 38.00 | 0.9601 | 33.64 | 0.9179 | 32.17 | 0.8990 | 32.19 | 0.9284 |
| | Cross-SRN | 2021 | 38.03 | 0.9606 | 33.62 | 0.9180 | 32.19 | **0.8997** | 32.28 | 0.9290 |
| | MBMFN | | **38.05** | 0.9599 | **33.78** | **0.9193** | **32.21** | 0.8996 | **32.44** | **0.9303** |
| ×3 | SRCNN | 2016 | 32.75 | 0.9090 | 29.28 | 0.8209 | 28.41 | 0.7863 | 26.24 | 0.7989 |
|  | VDSR | 2016 | 33.66 | 0.9213 | 29.77 | 0.8314 | 28.82 | 0.7976 | 27.14 | 0.8279 |
|  | DRCN | 2016 | 33.82 | 0.9226 | 29.76 | 0.8311 | 28.80 | 0.7963 | 27.15 | 0.8276 |
|  | MemNet | 2017 | 34.09 | 0.9248 | 30.00 | 0.8350 | 28.96 | 0.8001 | 27.56 | 0.8376 |
|  | CARN | 2018 | 34.29 | 0.9255 | 30.29 | 0.8407 | 29.06 | 0.8034 | 28.06 | 0.8493 |
|  | IMDN | 2019 | 34.36 | 0.9270 | 30.32 | 0.8417 | 29.09 | 0.8046 | 28.17 | 0.8519 |
|  | DNCL | 2019 | 33.95 | 0.9232 | 29.93 | 0.8340 | 28.91 | 0.7995 | 27.27 | 0.8326 |
|  | MRFN | 2019 | 34.21 | 0.9267 | 30.03 | 0.8363 | 28.99 | 0.8029 | 27.53 | 0.8389 |
|  | FilterNeL | 2020 | 34.08 | 0.9250 | 30.03 | 0.8370 | 28.95 | 0.8030 | 27.55 | 0.8380 |
|  | MADNet-$L_F$ | 2020 | 34.14 | 0.9251 | 30.20 | 0.8395 | 28.98 | 0.8023 | 27.78 | 0.8439 |
|  | RFDN | 2020 | 34.41 | 0.9273 | 30.34 | 0.8420 | 29.09 | 0.8050 | 28.21 | 0.8525 |
|  | CFSRCNN | 2020 | 34.24 | 0.9256 | 30.27 | 0.8410 | 29.03 | 0.8035 | 28.04 | 0.8496 |
|  | SeaNet-baseline | 2020 | 34.36 | **0.9280** | 30.34 | **0.8428** | 29.09 | 0.8053 | 28.17 | 0.8527 |
|  | SMSR | 2021 | 34.40 | 0.9270 | 30.33 | 0.8412 | 29.10 | 0.8050 | 28.25 | 0.8536 |
|  | Cross-SRN | 2021 | 34.43 | 0.9275 | 30.33 | 0.8417 | 29.09 | 0.8050 | 28.23 | 0.8535 |
|  | MBMFN |  | **34.52** | 0.9273 | **30.41** | 0.8426 | **29.12** | 0.8052 | **28.36** | **0.8553** |
| ×4 | SRCNN | 2016 | 30.48 | 0.8628 | 27.49 | 0.7503 | 26.90 | 0.7101 | 24.52 | 0.7221 |
|  | VDSR | 2016 | 31.35 | 0.8838 | 28.01 | 0.7674 | 27.29 | 0.7251 | 25.18 | 0.7524 |
|  | DRCN | 2016 | 31.53 | 0.8854 | 28.02 | 0.7670 | 27.23 | 0.7233 | 25.14 | 0.7510 |
|  | MemNet | 2017 | 31.74 | 0.8893 | 28.26 | 0.7723 | 27.40 | 0.7281 | 25.50 | 0.7630 |
|  | CARN | 2018 | 32.13 | 0.8937 | 28.60 | 0.7806 | 27.58 | 0.7349 | 26.07 | 0.7837 |
|  | IMDN | 2019 | 32.21 | 0.8948 | 28.58 | 0.7811 | 27.56 | 0.7353 | 26.04 | 0.7838 |
|  | DNCL | 2019 | 31.66 | 0.8871 | 28.23 | 0.7717 | 27.39 | 0.7282 | 25.36 | 0.7606 |
|  | MRFN | 2019 | 31.90 | 0.8916 | 28.31 | 0.7746 | 27.43 | 0.7309 | 25.46 | 0.7654 |
|  | FilterNeL | 2020 | 31.74 | 0.8900 | 28.27 | 0.7730 | 27.39 | 0.7290 | 25.53 | 0.7680 |
|  | MADNet-$L_F$ | 2020 | 32.01 | 0.8925 | 28.45 | 0.7781 | 27.47 | 0.7327 | 25.77 | 0.7751 |
|  | RFDN | 2020 | 32.24 | 0.8952 | 28.61 | 0.7819 | 27.57 | 0.7360 | 26.11 | 0.7858 |
|  | CFSRCNN | 2020 | 32.06 | 0.8920 | 28.57 | 0.7800 | 27.53 | 0.7333 | 26.03 | 0.7824 |
|  | SeaNet-baseline | 2020 | 32.18 | 0.8948 | 28.61 | 0.7822 | 27.57 | 0.7359 | 26.05 | 0.7896 |
|  | SMSR | 2021 | 32.12 | 0.8932 | 28.55 | 0.7808 | 27.55 | 0.7351 | 26.11 | 0.7868 |
|  | MSFIN | 2021 | 32.28 | **0.8957** | 28.57 | 0.7813 | 27.56 | 0.7358 | 26.13 | 0.7865 |
|  | Cross-SRN | 2021 | 32.24 | 0.8954 | 28.59 | 0.7817 | 27.58 | **0.7364** | 26.16 | 0.7881 |
|  | MBMFN |  | **32.31** | 0.8952 | **28.68** | **0.7829** | **27.60** | 0.7363 | **26.26** | **0.7899** |

From Table 4, we can see clearly that in the reconstruction tasks of×2, ×3, ×4, compared with other advanced algorithms such as Cross-SRN and SMSR, our MBMFN has produced excellent results on PSNR and SSIM. In addition, under the ×4 reconstruction task, for PSNR, MBMFN has higher 0.07dB, 0.09dB and 0.10dB than Cross-SRN on Set5, Set14 and Urban100 test sets respectively. On Set14 and Urban100, MBMFN is 0.0012 and 0.0018 higher than Cross-SRN on SSIM respectively. In particular, for the Urban100 test set containing a large number of detailed texture images, MBMFN is 0.034 higher than MSFIN and 0.031 higher than SMSR on SSIM, which shows that our MBMFN has great advantages in reconstructing high-frequency information such as edge texture.

In order to more intuitively verify that MBMFN has a high reconstruction ability for edge texture and other information, we visualize the reconstruction results under the ×4 reconstruction task, and the visual effect comparison is shown in figure 4. In the enlarged comparison of local images of Urban100_092 in Figure 4, the image reconstructed by CARN has serious line distortion and blur, and IMDN and RFDN also have different degrees of local line distortion. Although SMSR avoids line distortion, the reconstructed image has some blur. In contrast, our MBMFN restores more edge texture information of buildings, it is also closer to the original high-resolution image.

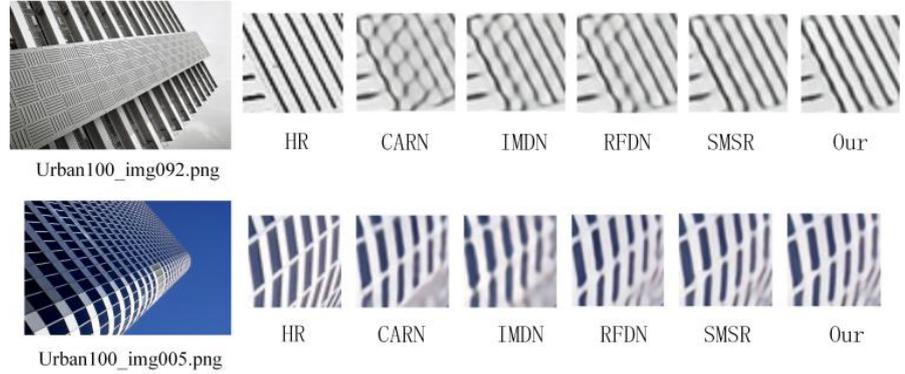

**Figure 4** Visual comparison on Urban100 datasets

In addition, in order to more intuitively compare the relationship between the performance of each algorithm and network parameters, under ×4 reconstruction task, we visualized the corresponding relationship between PSNR and parameters of some algorithms on Set5 test set. As shown in Figure 5, compared with some other advanced SR algorithms, MBMFN achieves the best performance under the condition that the number of network parameters is kept within a reasonable range. In particular, MBMFN achieves a good balance between parameters and performance, this makes it possible to apply it to small devices with limited memory and computing.

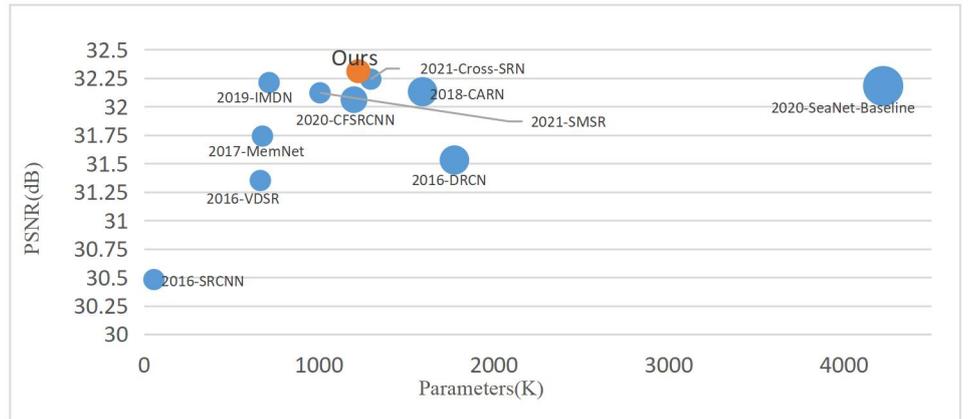

**Fig. 5** comparison of the relationship between model parameters and performance

## 5. Conclusions

In this paper, we propose a multi branch feature multiplexing fusion network with mixed multi-layer attention to realize SISR, and design a semi parallel multi branch feature multiplexing fusion attention block, which processes the local features extracted in different stages through multiple branches in parallel, and exchanges the feature information of different branches with the help of jump connection, It realizes the multiple utilization of local features and multi-stage feature fusion, expands the network receptive field, and avoids the problem that the network is difficult to train due to the deepening of the network. In addition, we designed a lightweight enhanced residual channel attention block to improve the sensitivity of the network to information rich channels, so as to learn more discriminative feature representation. We use the sum of global average and standard deviation of features to describe channel information more comprehensively, and use 1 × 1 convolution layer to directly model the relationship between channels, which avoids the loss of channel information caused by channel compression, this block can help the network recover more image details. In the reconstruction phase, we introduce attention mechanism and adopt distributed step-by-step up-sampling method to strengthen the network's attention and utilization of the high-frequency information in the features. Especially, few people study attention strategies during the reconstruction phase. We also use the weight sharing strategy to reduce the number of network parameters. In addition, as far as we know, we should be the first to combine the weight sharing strategy with the step-by-step up-sampling strategy in the reconstruction phase. A large number of experimental results show that our MBMFN has achieved excellent performance in both objective indicators and subjective vision.

**Funding:** This work was supported by the National Natural Science Foundation of China (Grant No. U1803261, Grant No. U1903213).

**Institutional Review Board Statement**: Not applicable.

**Informed Consent Statement**: Not applicable.

**Data Availability Statement**:

Our training set DIV2k datasets can be obtained from available online:

https://data.vision.ee.ethz.ch/cvl/DIV2K/

Set5, Set14, B100, Urban 100 can be obtained from Available online:

https://arxiv.org/abs/1909.11856

In addition, in order to facilitate learning and reproduce the experimental results, our code can be obtained in the following link:

https://github.com/Cai631/MBMFN

**Conflict of interest**: The authors declare no conflict of interest. The funders had no role in the design of the study; in the collection, analyses, or interpretation of data; in the writing of the manuscript, or in the decision to publish the results.

**References**

1. Guo Y, Chen J, Wang J, et al. Closed-loop matters: Dual regression networks for single image super-resolution[C]//Proceedings of the IEEE/CVF conference on computer vision and pattern recognition. 2020: 5407-5416.
2. Niu B, Wen W, Ren W, et al. Single image super-resolution via a holistic attention network[C]//European conference on computer vision. Springer, Cham, 2020: 191-207.
3. Luo X, Xie Y, Zhang Y, et al. Latticenet: Towards lightweight image super-resolution with lattice block[C]//European Conference on Computer Vision. Springer, Cham, 2020: 272-289.
4. Dong C, Loy C C, He K, et al. Image super-resolution using deep convolutional networks[J]. IEEE transactions on pattern analysis and machine intelligence, 2015, 38(2): 295-307.
5. Kim J, Lee J K, Lee K M. Accurate image super-resolution using very deep convolutional networks[C]//Proceedings of the IEEE conference on computer vision and pattern recognition. 2016: 1646-1654.

6. Lim B, Son S, Kim H, et al. Enhanced deep residual networks for single image super-resolution[C]//Proceedings of the IEEE conference on computer vision and pattern recognition workshops. 2017: 136-144.
7. Liu J, Zhang W, Tang Y, et al. Residual feature aggregation network for image super-resolution[C]//Proceedings of the IEEE/CVF conference on computer vision and pattern recognition. 2020: 2359-2368.
8. Liu J, Zhang W, Tang Y, et al. Residual feature aggregation network for image super-resolution[C]//Proceedings of the IEEE/CVF conference on computer vision and pattern recognition. 2020: 2359-2368.
9. Shi W, Caballero J, Huszár F, et al. Real-time single image and video super-resolution using an efficient sub-pixel convolutional neural network[C]//Proceedings of the IEEE conference on computer vision and pattern recognition. 2016: 1874-1883.
10. Hui Z, Wang X, Gao X. Fast and accurate single image super-resolution via information distillation network[C]//Proceedings of the IEEE conference on computer vision and pattern recognition. 2018: 723-731.
11. Hui Z, Gao X, Yang Y, et al. Lightweight image super-resolution with information multi-distillation network[C]//Proceedings of the 27th acm international conference on multimedia. 2019: 2024-2032.
12. Liu J, Tang J, Wu G. Residual feature distillation network for lightweight image super-resolution[C]//European Conference on Computer Vision. Springer, Cham, 2020: 41-55.
13. Wang Z, Gao G, Li J, et al. Lightweight Image Super-Resolution with Multi-scale Feature Interaction Network[C]//2021 IEEE International Conference on Multimedia and Expo (ICME). IEEE, 2021: 1-6.
14. Hu J, Shen L, Sun G. Squeeze-and-excitation networks[C]//Proceedings of the IEEE conference on computer vision and pattern recognition. 2018: 7132-7141.
15. Roy A G, Navab N, Wachinger C. Concurrent spatial and channel 'squeeze & excitation' in fully convolutional networks[C]//International conference on medical image computing and computer-assisted intervention. Springer, Cham, 2018: 421-429.
16. Wang Q, Wu B, Zhu P, et al. ECA-Net: Efficient Channel Attention for Deep Convolutional Neural Networks[C]// 2020 IEEE/CVF Conference on Computer Vision and Pattern Recognition (CVPR). IEEE, 2020:11534-11542.
17. Niu B, Wen W, Ren W, et al. Single image super-resolution via a holistic attention network[C]//European conference on computer vision. Springer, Cham, 2020: 191-207.
18. Liu J, Zhang W, Tang Y, et al. Residual feature aggregation network for image super-resolution[C]//Proceedings of the IEEE/CVF conference on computer vision and pattern recognition. 2020: 2359-2368.
19. Ledig C, Theis L, Huszár F, et al. Photo-realistic single image super-resolution using a generative adversarial network[C]//Proceedings of the IEEE conference on computer vision and pattern recognition. 2017: 4681-4690.
20. Lim B, Son S, Kim H, et al. Enhanced deep residual networks for single image super-resolution[C]//Proceedings of the IEEE conference on computer vision and pattern recognition workshops. 2017: 136-144.
21. Zhang Y, Li K, Li K, et al. Image super-resolution using very deep residual channel attention networks[C]//Proceedings of the European conference on computer vision (ECCV). 2018: 286-301.
22. Behjati P, Rodriguez P, Mehri A, et al. Hierarchical Residual Attention Network for Single Image Super-Resolution[J]. arXiv preprint arXiv:2012.04578, 2020.
23. Agustsson E, Timofte R. Ntire 2017 challenge on single image super-resolution: Dataset and study[C]//Proceedings of the IEEE conference on computer vision and pattern recognition workshops. 2017: 126-135.
24. Kim J, Lee J K, Lee K M. Deeply-recursive convolutional network for image super-resolution[C]//Proceedings of the IEEE conference on computer vision and pattern recognition. 2016: 1637-1645.
25. Tai Y, Yang J, Liu X, et al. Memnet: A persistent memory network for image restoration[C]//Proceedings of the IEEE international conference on computer vision. 2017: 4539-4547.

26. Ahn N, Kang B, Sohn K A. Fast, accurate, and lightweight super-resolution with cascading residual network[C]//Proceedings of the European conference on computer vision (ECCV). 2018: 252-268.
27. Xie C, Zeng W, Lu X. Fast single-image super-resolution via deep network with component learning[J]. IEEE Transactions on Circuits and Systems for Video Technology, 2018, 29(12): 3473-3486.
28. Li F, Bai H, Zhao Y. FilterNet: Adaptive Information Filtering Network for Accurate and Fast Image Super-Resolution[J]. IEEE Transactions on Circuits and Systems for Video Technology, 2020, 30(6):1511-1523.
29. Tian C, Xu Y, Zuo W, et al. Coarse-to-fine CNN for image super-resolution[J]. IEEE Transactions on Multimedia, 2020, 23: 1489-1502.
30. Fang F, Li J, Zeng T. Soft-edge assisted network for single image super-resolution[J]. IEEE Transactions on Image Processing, 2020, 29: 4656-4668.
31. Wang L, Dong X, Wang Y, et al. Exploring sparsity in image super-resolution for efficient inference[C]//Proceedings of the IEEE/CVF Conference on Computer Vision and Pattern Recognition. 2021: 4917-4926.
32. Liu Y, Jia Q, Fan X, et al. Cross-SRN: Structure-Preserving Super-Resolution Network with Cross Convolution[J]. IEEE Transactions on Circuits and Systems for Video Technology, 2021.
33. He Z, Cao Y, Du L, et al. MRFN: Multi-receptive-field network for fast and accurate single image super-resolution[J]. IEEE Transactions on Multimedia, 2019, 22(4): 1042-1054.
34. Lan R, Sun L, Liu Z, et al. MADNet: a fast and lightweight network for single-image super resolution[J]. IEEE transactions on cybernetics, 2020, 51(3): 1443-1453.